# Scaling Qualitative Analysis with Large Language Models: Development and Validation of the Smart Coder Framework for Human–LLM Collaborative Coding

Cheng Tang[1], Kun Wang[2], Yaxuan Yang[1], Shiyu Wang[1], Allan Cohen[1], Rachael Eriksen Brown[3], Chandra Hawley Orrill[2]

1 Department of Educational Psychology, University of Georgia, Athens, GA, USA

2 Rethink Learning Labs, Woodcliff Lake, NJ, USA

3 Department of Early Childhood/Elementary Education, Penn State Abington, Abington, PA, USA

Corresponding author: Shiyu Wang, swang44@uga.edu

# Scaling Qualitative Analysis with Large Language Models: Development and Validation of the Smart Coder Framework for Human–LLM Collaborative Coding

### Abstract

Current uses of large language models (LLMs) for qualitative coding often rely on ad hoc prompting with limited guidance for reliability and human oversight. To address these limitations, we propose the Smart Coder Framework, a three-phase approach structuring human–LLM collaborative coding through Instruction, Prompt Strategy, and Reliability Check. We evaluated the framework using 72 middle-school mathematics teacher interview transcripts and a human-consensus benchmark. A factorial design compared instructed versus uninstructed prompts, few-shot versus few-shot plus chain-of-thought prompting, and single-pass versus self-consistency coding. Twelve transcripts were used for prompt development and calibration, and 60 for the formal evaluation. In the results, instruction consistently improved agreement. Chain-of-thought further improved agreement under adequate instruction, and self-consistency produced smaller, theme-dependent gains but helped identify unstable decisions for human review. Performance also varied by transcript characteristics. Overall, the framework systematically structures and monitors LLM-assisted qualitative coding while preserving human authority over coding decisions.

## Introduction

Qualitative coding is the process of identifying meaningful segments of textual data and assigning them to concepts, categories, or themes through systematic interpretation (Braun and Clarke 2006; Saldaña 2021). In qualitative coding, these judgments are guided by a predefined codebook that specifies the constructs of interest and the criteria for applying each code. However, this process is labor-intensive because it requires repeated close reading, careful segmentation, and consistent interpretive judgment across large volumes of text (Gibbs 2018), thus making it difficult to scale (Campbell et al. 2013; O'Connor and Joffe 2020). In practice, this creates a scalability problem: as the volume of textual data grows, maintaining close, codebook-guided human interpretation becomes increasingly difficult, whereas automated approaches can process larger datasets but often provide less interpretive control and nuance.

Large language models (LLMs) offer a potential means of scaling this process. By following written instructions and examples included in the prompt, LLMs can apply predefined codebooks to textual data and generate coding decisions (Bijker et al. 2024; Tai et al. 2024). However, current applications often rely on ad hoc prompting practices instead of a systematic coding methodology. Researchers may specify the coding task, provide examples, request model reasoning, or repeat the task to assess stability, but they often use these choices separately, without clear guidance on their roles in the coding process. As a result, LLM-assisted coding may produce inconsistent decisions across prompts or repeated runs, while human input is introduced reactively through manual correction rather than systematically integrated into the coding procedure (Han et al. 2025). These limitations constrain both the reliability and scalability of LLM-assisted qualitative coding.

We identify three gaps that limit the reliable use of LLMs for qualitative coding. First, existing studies often evaluate individual prompting techniques in isolation, without situating these choices within a systematic methodology for designing and validating LLM-assisted coding. Second, reasoning strategies such as chain-of-thought (CoT) and stability checks such as self-consistency (SC) are often combined without distinguishing their different methodological functions, making it difficult to determine how each contributes to coding performance and reliability (Wei et al. 2022; Wang et al. 2023). Finally, evaluations typically emphasize aggregate performance, providing limited evidence about when coding accuracy varies across documents or coding themes. Such differences matter for identifying when greater human oversight may be needed (Liu et al. 2024).

To address these gaps, we propose the Smart Coder Framework, which treats LLM-assisted qualitative coding as a methodological design and validation problem. The framework structures human–LLM collaboration into three phases. Instruction aligns the model with human-defined rubrics. Prompt Strategy structures how the model evaluates textual evidence and generates a coding decision. Reliability Check evaluates the stability of coding decisions and identifies cases requiring further review. Human expertise remains central throughout the process, where researchers define the coding criteria, calibrate the coding procedure, establish the benchmark, and retain authority over interpretation and adjudication.

We empirically evaluate the framework using mathematics-education interview transcripts through a $2 \times 2 \times 2$ factorial design that manipulates design choices corresponding to the three phases: instructed versus uninstructed coding, few-shot versus few-shot plus CoT prompting, and single-pass versus SC coding. We address three research questions:

RQ1: How does explicit task instruction affect LLM coding performance relative to uninstructed coding?

RQ2: How do reasoning optimization through CoT and reliability checking through SC independently and jointly relate to coding performance?

RQ3: How does coding performance vary across transcript characteristics and coding themes?

This study contributes an empirically evaluated methodology for designing and validating human–LLM collaborative deductive coding. The Smart Coder Framework specifies how researchers can align an LLM with a human-developed coding rubric, guide how it uses textual evidence, evaluate the stability of its decisions, and compare its codes with expert judgment. In doing so, the framework provides a transparent structure for scaling qualitative coding while retaining human control over code definition, interpretation, and adjudication.

## Methodological Rationale for the Smart Coder Framework

LLM-assisted qualitative coding involves several researcher-controlled design choices, including how the coding task is specified, how the model is prompted to use textual evidence, and how variability in model outputs is handled (Zhao et al. 2026). Although researchers typically treat the pretrained model as fixed, they can modify task instructions, in-context examples, prompting strategies, and generation settings such as temperature and top-p (Holtzman et al. 2020; Liu et al. 2023; Wei et al. 2022). These choices are often treated as technical implementation decisions. For qualitative coding, however, they also have methodological consequences because they influence how human-developed coding criteria are communicated to the model, how coding decisions are produced, and how stable those decisions are across repeated runs.

Recent studies demonstrate that LLMs can support qualitative coding, including broad thematic analysis (Morgan 2023) and classroom-dialogue coding with substantial agreement with expert judgments and reduced coding time (Long et al. 2024). At the same time, performance varies across coding tasks and levels of interpretive complexity (Bijker et al. 2024; Tai et al. 2024). This variability makes task alignment particularly important for deductive coding, in which judgments must be made according to predefined categories and decision criteria (Guba 1979; O'Connor and Joffe 2020). Iterative prompt refinement can improve this alignment (Jeong et al. 2025), but such refinement is often treated as an informal development step rather than an explicit methodological process for establishing correspondence between the human coding rubric and the model's application of that rubric.

Once task alignment has been established, a second methodological decision concerns how the model is asked to translate textual evidence into a coding judgment. Few-shot prompting provides in-context examples that illustrate how the coding criteria should be applied (Brown et al. 2020), whereas CoT prompting can elicit an explicit sequence of intermediate steps linking textual evidence to a code (Kojima et al. 2022). Studies combining examples with explicit rationales have reported strong coding performance (Amara et al. 2024; Hamerlik et al. 2024), but these strategies are often bundled together, making their distinct contributions difficult to evaluate. More importantly, strategies intended to structure the coding process serve a different methodological function from procedures intended to evaluate whether the resulting decision is stable.

This distinction motivates a separate reliability check. SC generates repeated sampled responses under the same coding condition and compares or aggregates the resulting decisions (Wang et al. 2023). Repeated agreement can provide information about decision stability, but consistency alone does not establish correctness relative to a human benchmark (Bartsch et al. 2023). SC is therefore more appropriately viewed as a diagnostic for identifying unstable cases than as a substitute for validation or human judgment. In addition, LLM performance may vary across source documents because models can be sensitive to input length and to how relevant evidence is distributed within longer contexts (Han et al. 2025; Hsieh et al. 2024; Liu et al. 2024). In summary, these considerations motivate a methodological framework that separates task alignment, evidence use in code selection, and assessment of decision stability, while maintaining human oversight of rubric definition, calibration, and adjudication.

## The Smart Coder Framework

The Smart Coder Framework structures LLM-assisted qualitative coding into three methodological phases: Instruction, Prompt Strategy, and Reliability Check. Each phase addresses a distinct function in the coding process. The Instruction Phase focuses on aligning the model with a human-developed coding rubric. The Prompt Strategy Phase specifies how the model is asked to move from textual evidence to a coding decision. The Reliability Check Phase evaluates the stability of codes across repeated runs.

A key feature of the framework is the distinction between calibration and formal evaluation. Calibration precedes formal evaluation and asks whether the coding procedure is sufficiently aligned with the human-developed rubric to proceed. Formal evaluation then asks how well the calibrated procedure performs on data not used for development. Both rely on comparison with human judgment, but they serve different methodological purposes. Figure 1 summarizes the overall structure of the Smart Coder Framework.

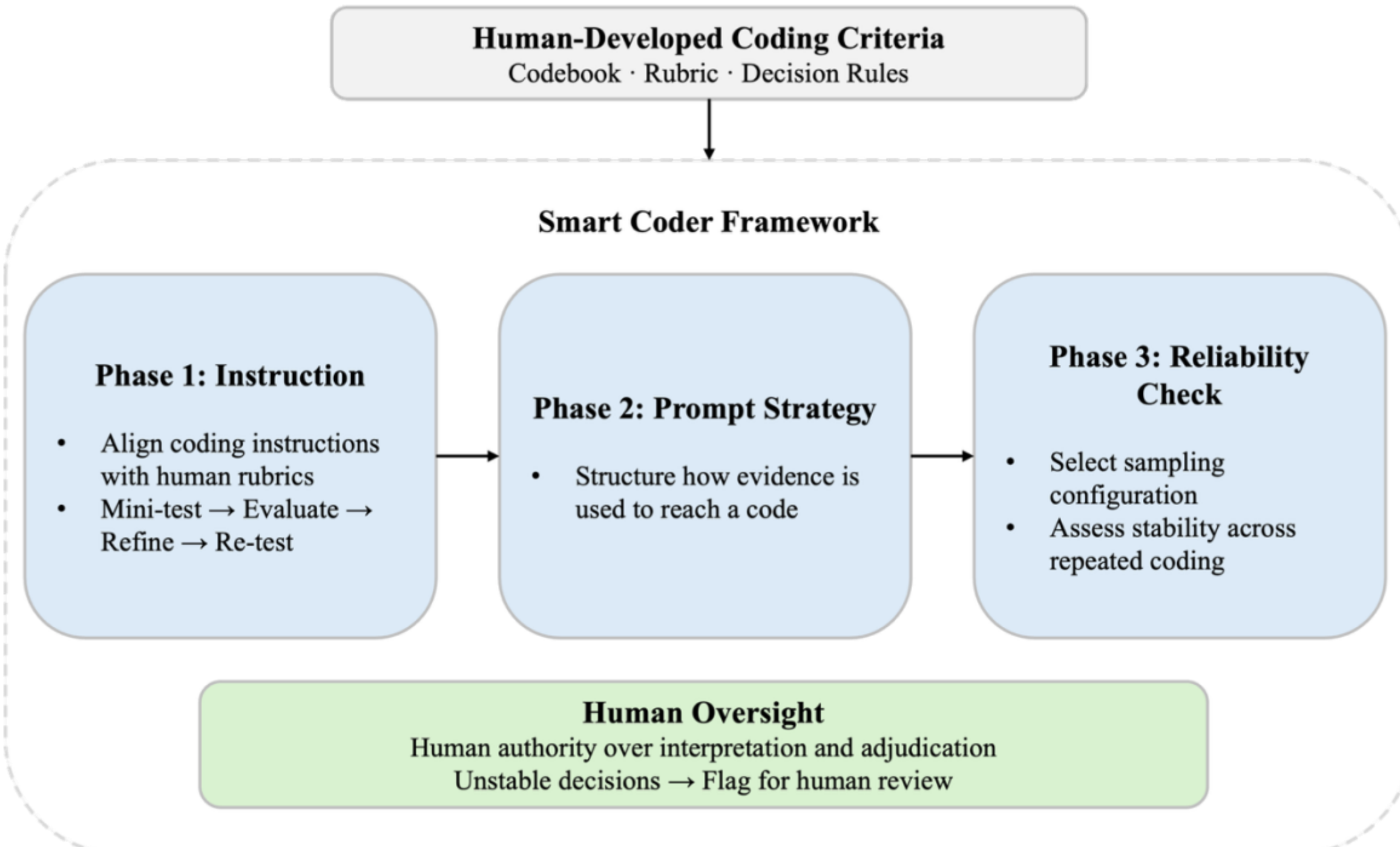

**Figure 1.** Workflow of the Smart Coder Framework.

**Phase 1: Instruction**

The Instruction Phase is a human-guided alignment process. Rather than fine-tuning the pretrained model, researchers iteratively clarify rubric wording, decision rules, and coding examples so that the instructions presented to the LLM reflect the intended human coding criteria.

Alignment is assessed through a mini-test using human-labeled transcripts that are excluded from the formal comparison set. Model-generated codes are compared with the human benchmark to determine whether the current instructions are sufficiently aligned to advance. In this study, agreement was evaluated using Cohen's kappa (Cohen 1960), and a kappa greater than 0.61, the conventional boundary for substantial agreement (Landis and Koch 1977), was used as the advancement criterion. When a candidate prompt fell below the criterion, researchers examined disagreements, clarified distinctions between adjacent scores, adjusted examples to cover edge cases and themes, and re-ran the mini-test. The mini-test therefore serves as the calibration step, rather than as evidence from the formal evaluation.

**Phase 2: Prompt Strategy**

Once the coding instructions align with the human rubric, the Prompt Strategy Phase determines how the model uses evidence to select a code. Few-shot prompting, using the coding rubric and examples, serves as the baseline. The CoT condition adds four ordered reasoning steps: identifying theme-relevant text, comparing that evidence with the conceptual definition, assessing the strength or completeness of the evidence, and mapping the judgment to the rubric score. In this study, CoT provides an explicit record of the evidence to code response structure for each coding decision. This phase distinguishes task alignment, established during the Instruction Phase, from the response strategy used to produce a coding decision.

**Phase 3: Reliability Check**

The Reliability Check Phase evaluates the stability of coding outputs across repeated model runs. Under SC, the same transcript is submitted repeatedly under the same prompt condition, and the resulting codes are aggregated by majority voting. Comparing repeated outputs reduces reliance on a single sampled response and provides a vote distribution that can be used to identify less stable coding decisions for human review. The sampling configuration was calibrated on the development set by varying iteration count (N), temperature, and top-p.

## Calibration and Formal Evaluation

Evaluation serves two distinct purposes within the Smart Coder Framework. First, during calibration, the mini-test provides a readiness check: model-generated codes are compared with the human benchmark to determine whether the coding procedure is sufficiently aligned with the rubric to proceed to formal evaluation. Second, after calibration, formal evaluation assesses the performance of the resulting coding procedure using transcripts that were not used in the mini-test.

Because no single agreement index fully characterizes coding performance, formal evaluation can use complementary measures appropriate to the coding task. In the present study, we use Percent Agreement (PA), Cohen's kappa, Gwet's AC1, and Document-Level Exact Match. PA summarizes raw agreement. Kappa adjusts for chance agreement but can be sensitive to skewed category distributions, whereas Gwet's AC1 provides a complementary chance-corrected measure that is more stable under category imbalance (Cohen 1960; Gwet 2008). Document-Level Exact Match assesses whether all theme-level codes for a transcript agree with the human benchmark. The specific indices may differ across coding tasks, but the methodological distinction remains the same: the Instruction-phase mini-test is used to calibrate the coding procedure, whereas formal evaluation is used to assess the performance of the calibrated procedure on separate data.

## Empirical Evaluation

### Coding Task and Data

The coding task required assigning one ordinal score to each transcript for each of four mathematical themes: Referent Unit, Covariance, Invariance/Constant, and Quantity. Referent Unit concerns the conceptual whole to which a fraction or value refers. Covariance refers to the coordinated change of two quantities within a rational relationship. Invariance/Constant is a relationship or value that remains unchanged under transformation. Quantity is a value understood through its magnitude, unit, and referent. Scores were 0 (no evidence), 1 (partial or implicit evidence), or 2 (explicit, well-developed evidence), as shown in Table 1.

The corpus comprised 72 interview transcripts from middle-school mathematics teachers in New Jersey and Florida. Interviews followed an expert-designed semi-structured protocol of 11 primary questions and 36 sub-questions about proportion and fractions, eliciting evidence of teachers' mathematical knowledge (Shulman 1986). Each transcript served as the unit of analysis.

**Table 1.** The Three-Level Coding Rubric

| Coding | Meaning | Operational Definition |
|---|---|---|
| 0 | No evidence | The theme is absent or incorrectly applied |
| 1 | Partial evidence | The theme is present but implicit or incomplete |
| 2 | Explicit evidence | The theme is clearly articulated with supporting logic |

Two qualitative experts collaboratively refined the rubric and its decision boundaries, independently reviewed the transcripts and scored each theme, and resolved disagreements through discussion until consensus. Their consensus scores formed the benchmark. We used 12 transcripts as the development set. Five were reserved for the mini-test used to evaluate candidate prompts and sampling configurations, while the other seven served as the pool of candidate few-shot examples. The remaining 60 were used as the formal comparison set.

## Document-Level Features

For RQ3, we characterize each transcript with Total Token Count, Useful Token Count, Useful Token Ratio, and Tokens per Sentence. Useful tokens remain after removing punctuation, extraneous whitespace, and stopwords and applying lemmatization. The ratio divides useful by total tokens. Tokens per Sentence reflects the average sentence length. These descriptive features allow comparison across transcripts with different lengths and lexical content distributions.

## Experimental Design and Configuration

We used a structured 2 x 2 x 2 comparison with three binary factors: Instruction Status (instructed or uninstructed), Prompt Strategy (few-shot or few-shot plus CoT), and Reliability Check (single-pass or SC), yielding eight conditions. Uninstructed conditions used the initial prompt before iterative instructional refinement. Instructed conditions used the refined five-example prompt retained after development-set calibration. The CoT conditions added the four-step reasoning process. Single-pass conditions generated one response, whereas SC conditions generated 10 responses for majority voting. The same 60 formal-comparison transcripts were coded across all conditions, and the condition results are paired descriptive comparisons. The full condition matrix appears in Supplemental Table S1.

GPT-4.1 (OpenAI 2025) served as the base model in all conditions. For SC calibration, we varied one parameter at a time across the candidate values reported in Supplemental Table S2.

Because GPT-4.1 was used in every condition, differences across conditions cannot be attributed to changes in the underlying model. These comparisons remain descriptive and do not establish causal effects or generalizability to other LLMs.

## Results

Results across the eight experimental conditions are organized by the framework phases and then by document features. Prompt and sampling calibration results from the development set are reported in Supplemental Tables S3 and S4. Overall, the clearest performance differences were associated with instruction, whereas patterns for CoT, SC, and document characteristics were smaller or more theme-specific.

### *RQ1: Instruction and Coding Performance*

Candidate instructed prompts with three to seven examples were evaluated in the mini-test. Four produced kappa values above 0.61 and identical scores on all agreement measures, as reported in Supplemental Table S3. We retained the five-example prompt because it offered broader coverage than the shorter alternative while remaining more concise than the longer versions.

Across the eight formal conditions, the largest observed differences separated instructed from uninstructed prompts, as shown in Table 2. For example, adding CoT to the uninstructed prompt reduced PA from 0.69 to 0.63, kappa from 0.43 to 0.36, and Document-Level Exact Match from 0.28 to 0.27. In this dataset, advanced response structures did not compensate for an unrefined prompt, supporting the workflow decision to complete instructional calibration before evaluating later phases.

**Table 2.** Comparison Between Instructed and Uninstructed Prompts

| Evaluation Metric | Prompt Configuration | Instructed | Uninstructed |
|---|---|---|---|
| Percent Agreement (PA) | Few-Shot | 0.76 | 0.69 |
| | Few-Shot + CoT | 0.80 | 0.63 |
| | Few-Shot + SC | 0.78 | 0.66 |
| | Few-Shot + CoT + SC | 0.81 | 0.65 |
| Cohen's Kappa | Few-Shot | 0.49 | 0.43 |
| | Few-Shot + CoT | 0.60 | 0.36 |
| | Few-Shot + SC | 0.53 | 0.37 |
| Cohen's Kappa (continued) | Few-Shot + CoT + SC | 0.61 | 0.38 |
| Gwet's AC1 | Few-Shot | 0.87 | 0.58 |
| | Few-Shot + CoT | 0.89 | 0.50 |

| Evaluation Metric | Prompt Configuration | Instructed | Uninstructed |
|---|---|---|---|
| | Few-Shot + SC | 0.89 | 0.54 |
| | Few-Shot + CoT + SC | 0.90 | 0.52 |
| Document Exact Match | Few-Shot | 0.33 | 0.28 |
| | Few-Shot + CoT | 0.47 | 0.27 |
| | Few-Shot + SC | 0.38 | 0.25 |
| | Few-Shot + CoT + SC | 0.45 | 0.28 |

*RQ2: Prompt Strategy, Self-Consistency, and Coding Performance*

Within instructed conditions, adding CoT was associated with higher overall performance: Document-Level Exact Match increased from 0.33 to 0.47, PA from 0.76 to 0.80, and kappa from 0.49 to 0.60. The overall tabulations are in Supplemental Table S5. The clearest theme-level difference was for Quantity, where PA increased from 0.55 to 0.67 and kappa from 0.24 to 0.39. CoT did not improve Covariance, indicating that the observed association was not uniform across themes.

All tested sampling configurations in the mini-test yielded identical agreement values, as reported in Supplemental Table S4. We therefore retained the SC configuration used: N = 10, temperature = 0.5, and top-p = 1.0 as a balanced configuration for the formal evaluation, considering the number of repeated responses needed to assess coding stability, the degree of sampling variation, and computational cost.

SC produced smaller overall differences, as shown in Supplemental Table S6. Its clearest theme-level pattern was for Covariance: under the instructed few-shot prompt, adding SC increased PA from 0.93 to 0.97 and kappa from 0.20 to 0.49. Repeated runs also produced vote distributions that can be inspected as a stability diagnostic. In these data, CoT was associated with improvement across several agreement measures, whereas the clearest change under SC occurred for Covariance.

*RQ3: Coding Performance Across Transcript Characteristics and Themes*

Document-Level Exact Match was compared with the four transcript features, as indicated in Supplemental Table S7. Across all eight conditions, correctly coded transcripts were shorter on average than incorrectly coded transcripts: mean Total Token Count differences ranged from -92.43 to -40.98, illustrated in Supplemental Figure S1, and Useful Token Count differences ranged from -34.00 to -13.15. Useful Token Ratio differed little. Correctly coded transcripts also tended to have more tokens per sentence, an observed association with average sentence length rather than evidence about syntactic complexity, shown in Supplemental Figure S2.

Patterns varied by theme, as indicated in Supplemental Table S8. Correct Referent Unit and Quantity scores were generally associated with shorter transcripts, whereas correctly coded Covariance cases under instructed CoT conditions had higher total and useful token counts. The total token pattern is illustrated in Supplemental Figure S3. These descriptive results caution against applying one preprocessing rule across themes: some relational concepts may depend on evidence distributed across longer responses.

## Discussion and Conclusion

This study shifts attention from whether LLMs can perform qualitative coding to the workflow conditions under which their codes align with a human-consensus benchmark. In this empirical setting, the largest performance differences were associated with prompt instruction, while structured response generation and repeated sampling had more conditional roles.

Prompt calibration produced the largest differences among the conditions examined. Instructed prompts consistently exceeded corresponding uninstructed prompts, and adding CoT or SC before prompt refinement sometimes reduced agreement. The result resembles human coder preparation: a codebook becomes operational through practice on decision boundaries and edge cases. The development set should therefore be treated as a calibration resource, not as independent evidence of generalization.

CoT and SC were not interchangeable. Requesting an explicit explanation of how the evidence supports the rubric score was associated with higher overall agreement under the instructed prompt, particularly for Quantity, but it did not improve every theme. Such rationales are useful as reviewable artifacts, not as guaranteed accounts of model cognition. SC yielded smaller overall changes but offered repeated-vote information and a marked descriptive gain for Covariance. Its value is therefore both aggregative and diagnostic.

Performance also varied with transcript characteristics. Shorter transcripts were generally associated with exact matches, but Covariance sometimes showed the opposite pattern, consistent with relational evidence being distributed across context. Tokens per Sentence should be interpreted as average sentence length rather than syntactic density. These results support theme-sensitive preprocessing and human review rather than a universal length-reduction rule.

The agreement results also illustrate why multiple indices are needed. In the instructed few-shot Covariance condition, PA was 0.93, and Gwet's AC1 was 0.97 while kappa was 0.20. Uneven category distributions can suppress kappa despite high observed agreement, so PA, kappa, AC1, and exact match should be interpreted together rather than as interchangeable measures.

This study has several limitations. The analysis used one mathematics-education dataset and one LLM, so the numerical patterns should not be generalized across domains or models. Prompt selection and sampling calibration were evaluated on the five mini-test transcripts, making the agreement values developmental rather than confirmatory. The same 60 transcripts were coded under every condition, and no inferential model of the paired observations was reported. Consequently, the condition contrasts are descriptive. Finally, the human consensus benchmark also reflects one rubric and the judgments of the participating experts.

The Smart Coder Framework offers a structured approach to LLM-assisted deductive coding that keeps human experts responsible for rubric definition, calibration, interpretation, and adjudication. Its methodological value lies in making those responsibilities and checkpoints explicit, not in replacing qualitative judgment with model output.

## Author Contributions

Cheng Tang: Conceptualization, Methodology, Software, Validation, Formal Analysis, Investigation, Data Curation, Writing – Original Draft, Writing – Review & Editing, Visualization.

Kun Wang: Conceptualization, Validation, Investigation, Resources, Data Curation, Writing – Review & Editing.

Yaxuan Yang: Conceptualization, Methodology, Validation, Data Curation, Writing – Review & Editing.

Shiyu Wang: Conceptualization, Methodology, Validation, Formal Analysis, Writing – Review & Editing, Supervision, Project Administration.

Allan Cohen: Conceptualization, Methodology, Writing – Review & Editing, Supervision.

Rachael Eriksen Brown: Conceptualization, Validation, Investigation, Resources, Data Curation, Writing – Review & Editing.

Chandra Hawley Orrill: Conceptualization, Validation, Investigation, Resources, Data Curation, Writing – Review & Editing.

## Statements and Declarations

Declaration of Conflicting Interest: There are no conflicts of interest.

Data Availability: Not applicable.

## Supplemental Materials

This supplement contains eight tables and three figures referenced in the manuscript. Together with the exhibits retained in the manuscript, it includes every table and figure from the original manuscript and appendix.

**Table S1.** Eight Conditions of the 2 × 2 × 2 Experimental Design

| Condition | Factor 1: Instruction | Factor 2: Prompt Strategy | Factor 3: Reliability Check |
|---|---|---|---|
| 1 | Uninstructed | Few-Shot | Single-Pass |
| 2 | Uninstructed | Few-Shot | SC |
| 3 | Uninstructed | Few-Shot+CoT | Single-Pass |
| 4 | Uninstructed | Few-Shot+CoT | SC |
| 5 | Instructed | Few-Shot | Single-Pass |
| 6 | Instructed | Few-Shot | SC |
| 7 | Instructed | Few-Shot+CoT | Single-Pass |
| 8 | Instructed | Few-Shot+CoT | SC |

**Table S2.** Self-Consistency Calibration Components and Tested Values

| Calibration Component | Parameter | Tested Values |
|---|---|---|
| Iteration count | N | 5, 10, 20 |
| Sampling variability | Temperature | 0.3, 0.5, 0.7, 0.9 |
| Nucleus sampling | top_p | 0.90, 0.95, 1.00 |

**Table S3.** Mini-Test Prompt Results

| Condition | PA | Cohen's Kappa | Gwet_AC1 | Doc_Match |
|---|---|---|---|---|
| 3-example | 0.90 | 0.74 | 0.88 | 0.60 |
| 4-example | 0.80 | 0.55 | 0.75 | 0.20 |
| 5-example | 0.90 | 0.74 | 0.88 | 0.60 |
| 6-example | 0.90 | 0.74 | 0.88 | 0.60 |
| 7-example | 0.90 | 0.74 | 0.88 | 0.60 |

**Table S4.** Self-Consistency Hyperparameter Tuning Results

| Condition | N | Temperature | Top_p | PA | Cohen's Kappa | Gwet's AC1 | Doc Match |
|---|---|---|---|---|---|---|---|
| 1 | 5 | 0.5 | 1.00 | 0.80 | 0.52 | 0.75 | 0.20 |
| 2 | 10 | 0.5 | 1.00 | 0.80 | 0.52 | 0.75 | 0.20 |
| 3 | 20 | 0.5 | 1.00 | 0.80 | 0.52 | 0.75 | 0.20 |
| 4 | 10 | 0.3 | 1.00 | 0.80 | 0.52 | 0.75 | 0.20 |
| 5 | 10 | 0.7 | 1.00 | 0.80 | 0.52 | 0.75 | 0.20 |
| 6 | 10 | 0.9 | 1.00 | 0.80 | 0.52 | 0.75 | 0.20 |
| 7 | 10 | 0.5 | 0.95 | 0.80 | 0.52 | 0.75 | 0.20 |
| 8 | 10 | 0.5 | 0.90 | 0.80 | 0.52 | 0.75 | 0.20 |

**Table S5.** Effect of Chain-of-Thought Across Instructed and Uninstructed Prompts

| Condition Context | Prompt Strategy | PA | Cohen's Kappa | Gwet's AC1 | Doc Exact Match |
|---|---|---|---|---|---|
| Instructed | Few-Shot (Baseline) | 0.76 | 0.49 | 0.87 | 0.33 |
| | Few-Shot + CoT | 0.8 | 0.6 | 0.89 | 0.47 |
| Uninstructed | Few-Shot (Baseline) | 0.69 | 0.43 | 0.58 | 0.28 |
| | Few-Shot + CoT | 0.63 | 0.36 | 0.5 | 0.27 |

**Table S6.** Effect of Self-Consistency Across Instructed and Uninstructed Prompts

| Condition Context | Prompt Strategy | PA | Cohen's Kappa | Gwet's AC1 | Doc Exact Match |
|---|---|---|---|---|---|
| Instructed | Few-Shot | 0.76 | 0.49 | 0.87 | 0.33 |
| | Few-Shot + SC | 0.78 | 0.53 | 0.89 | 0.38 |
| | Few-Shot + CoT | 0.80 | 0.60 | 0.89 | 0.47 |
| | Few-Shot + CoT + SC | 0.81 | 0.61 | 0.90 | 0.45 |
| Uninstructed | Few-Shot | 0.69 | 0.43 | 0.58 | 0.28 |

| Condition Context | Prompt Strategy | PA | Cohen's Kappa | Gwet's AC1 | Doc Exact Match |
|---|---|---|---|---|---|
| | Few-Shot + SC | 0.66 | 0.37 | 0.54 | 0.25 |
| | Few-Shot + CoT | 0.63 | 0.36 | 0.50 | 0.27 |
| | Few-Shot + CoT + SC | 0.65 | 0.38 | 0.52 | 0.28 |

**Table S7.** Document-Level Feature Characteristics

| Correct | Feature | N | Mean | SD |
|---|---|---|---|---|
| | **Uninstructed Few-Shot** | | | |
| 0 | tokens_per_sentence | 43 | 12.70 | 7.16 |
| 0 | total_token_count | 43 | 194.58 | 116.68 |
| 0 | useful/total | 43 | 0.36 | 0.05 |
| 0 | useful_token_count | 43 | 67.91 | 41.38 |
| 1 | tokens_per_sentence | 17 | 12.88 | 7.18 |
| 1 | total_token_count | 17 | 105.94 | 64.08 |
| 1 | useful/total | 17 | 0.34 | 0.06 |
| 1 | useful_token_count | 17 | 35.88 | 23.87 |
| | **Uninstructed Few-Shot-SC** | | | |
| 0 | tokens_per_sentence | 45 | 12.48 | 7.08 |
| 0 | total_token_count | 45 | 190.18 | 113.79 |
| 0 | useful/total | 45 | 0.36 | 0.05 |
| 0 | useful_token_count | 45 | 66.91 | 40.88 |
| 1 | tokens_per_sentence | 15 | 13.57 | 7.37 |
| 1 | total_token_count | 15 | 107.33 | 78.53 |
| 1 | useful/total | 15 | 0.34 | 0.06 |
| 1 | useful_token_count | 15 | 34.60 | 24.29 |
| | **Uninstructed Few-Shot-CoT** | | | |
| 0 | tokens_per_sentence | 44 | 12.23 | 7.11 |
| 0 | total_token_count | 44 | 193.55 | 117.07 |
| 0 | useful/total | 44 | 0.36 | 0.05 |
| 0 | useful_token_count | 44 | 67.73 | 41.85 |
| 1 | tokens_per_sentence | 16 | 14.18 | 7.12 |
| 1 | total_token_count | 16 | 103.25 | 56.75 |
| 1 | useful/total | 16 | 0.34 | 0.06 |
| 1 | useful_token_count | 16 | 34.38 | 18.57 |

**Table S7 (continued).** Document-Level Feature Characteristics

| Correct | Feature | N | Mean | SD |
|---|---|---|---|---|
| | **Uninstructed Few-shot-CoT-SC** | | | |
| 0 | tokens_per_sentence | 43 | 12.36 | 7.14 |
| 0 | total_token_count | 43 | 195.33 | 117.85 |
| 0 | useful/total | 43 | 0.36 | 0.05 |
| 0 | useful_token_count | 43 | 68.47 | 42.05 |
| 1 | tokens_per_sentence | 17 | 13.73 | 7.14 |
| 1 | total_token_count | 17 | 104.06 | 55.04 |
| 1 | useful/total | 17 | 0.34 | 0.06 |
| 1 | useful_token_count | 17 | 34.47 | 17.98 |
| | **Instructed Few-Shot** | | | |
| 0 | tokens_per_sentence | 40 | 11.96 | 6.21 |
| 0 | total_token_count | 40 | 200.28 | 118.18 |
| 0 | useful/total | 40 | 0.35 | 0.05 |
| 0 | useful_token_count | 40 | 69.28 | 41.96 |
| 1 | tokens_per_sentence | 20 | 14.34 | 8.59 |
| 1 | total_token_count | 20 | 107.85 | 62.68 |
| 1 | useful/total | 20 | 0.35 | 0.06 |
| 1 | useful_token_count | 20 | 37.95 | 24.89 |
| | **Instructed Few-Shot-SC** | | | |
| 0 | tokens_per_sentence | 37 | 11.86 | 6.44 |
| 0 | total_token_count | 37 | 196.78 | 123.10 |
| 0 | useful/total | 37 | 0.35 | 0.04 |
| 0 | useful_token_count | 37 | 68.24 | 43.70 |
| 1 | tokens_per_sentence | 23 | 14.19 | 8.01 |
| 1 | total_token_count | 23 | 125.52 | 72.54 |
| 1 | useful/total | 23 | 0.35 | 0.06 |
| 1 | useful_token_count | 23 | 43.70 | 27.17 |

**Table S7 (continued).** Document-Level Feature Characteristics

| Correct | Feature | N | Mean | SD |
|---|---|---|---|---|
| | **Instructed Few-Shot-CoT** | | | |
| 0 | tokens_per_sentence | 32 | 12.29 | 6.91 |
| 0 | total_token_count | 32 | 189.88 | 113.98 |
| 0 | useful/total | 32 | 0.35 | 0.05 |
| 0 | useful_token_count | 32 | 64.97 | 39.76 |
| 1 | tokens_per_sentence | 28 | 13.28 | 7.42 |
| 1 | total_token_count | 28 | 146.14 | 105.76 |
| 1 | useful/total | 28 | 0.35 | 0.06 |
| 1 | useful_token_count | 28 | 51.82 | 39.41 |
| | **Instructed Few-shot-CoT-SC** | | | |
| 0 | tokens_per_sentence | 33 | 12.49 | 6.84 |
| 0 | total_token_count | 33 | 187.91 | 115.49 |
| 0 | useful/total | 33 | 0.35 | 0.05 |
| 0 | useful_token_count | 33 | 64.79 | 40.38 |
| 1 | tokens_per_sentence | 27 | 13.08 | 7.54 |
| 1 | total_token_count | 27 | 146.93 | 104.06 |
| 1 | useful/total | 27 | 0.35 | 0.06 |
| 1 | useful_token_count | 27 | 51.56 | 38.61 |

**Table S8.** Theme-Level Feature Characteristics

| Theme | Correct | Feature | N | Mean | SD |
|---|---|---|---|---|---|
| | | **Uninstructed Few-Shot** | | | |
| Referent unit | 0 | tokens_per_sentence | 14 | 9.74 | 2.33 |
| Referent unit | 0 | total_token_count | 14 | 181.14 | 105.74 |
| Referent unit | 0 | useful/total | 14 | 0.35 | 0.04 |
| Referent unit | 0 | useful_token_count | 14 | 63.14 | 38.73 |
| Referent unit | 1 | tokens_per_sentence | 46 | 13.67 | 7.81 |
| Referent unit | 1 | total_token_count | 46 | 165.91 | 114.06 |
| Referent unit | 1 | useful/total | 46 | 0.35 | 0.06 |
| Referent unit | 1 | useful_token_count | 46 | 57.52 | 40.47 |
| Quantity | 0 | tokens_per_sentence | 23 | 13.60 | 7.65 |
| Quantity | 0 | total_token_count | 23 | 208.13 | 126.24 |
| Quantity | 0 | useful/total | 23 | 0.35 | 0.05 |
| Quantity | 0 | useful_token_count | 23 | 72.78 | 46.36 |
| Quantity | 1 | tokens_per_sentence | 37 | 12.22 | 6.80 |
| Quantity | 1 | total_token_count | 37 | 145.43 | 95.30 |
| Quantity | 1 | useful/total | 37 | 0.35 | 0.06 |
| Quantity | 1 | useful_token_count | 37 | 50.16 | 32.91 |
| Invariance/Constant | 0 | tokens_per_sentence | 18 | 13.04 | 8.24 |
| Invariance/Constant | 0 | total_token_count | 18 | 230.61 | 117.29 |
| Invariance/Constant | 0 | useful/total | 18 | 0.36 | 0.06 |
| Invariance/Constant | 0 | useful_token_count | 18 | 81.61 | 41.41 |
| Invariance/Constant | 1 | tokens_per_sentence | 42 | 12.63 | 6.67 |
| Invariance/Constant | 1 | total_token_count | 42 | 143.26 | 99.17 |
| Invariance/Constant | 1 | useful/total | 42 | 0.35 | 0.05 |
| Invariance/Constant | 1 | useful_token_count | 42 | 49.07 | 35.28 |
| Covariance | 0 | tokens_per_sentence | 19 | 14.35 | 10.02 |
| Covariance | 0 | total_token_count | 19 | 226.11 | 108.01 |
| Covariance | 0 | useful/total | 19 | 0.35 | 0.04 |
| Covariance | 0 | useful_token_count | 19 | 78.26 | 35.45 |
| Covariance | 1 | tokens_per_sentence | 41 | 12.01 | 5.25 |
| Covariance | 1 | total_token_count | 41 | 143.22 | 104.17 |
| Covariance | 1 | useful/total | 41 | 0.35 | 0.06 |
| Covariance | 1 | useful_token_count | 41 | 49.83 | 38.86 |

**Table S8 (continued).** Theme-Level Feature Characteristics

| Theme | Correct | Feature | N | Mean | SD |
|---|---|---|---|---|---|
| | | **Uninstructed Few-Shot-SC** | | | |
| Referent unit | 0 | tokens_per_sentence | 17 | 9.55 | 2.31 |
| Referent unit | 0 | total_token_count | 17 | 187.65 | 101.37 |
| Referent unit | 0 | useful/total | 17 | 0.36 | 0.04 |
| Referent unit | 0 | useful_token_count | 17 | 66.24 | 37.49 |
| Referent unit | 1 | tokens_per_sentence | 43 | 14.02 | 7.95 |
| Referent unit | 1 | total_token_count | 43 | 162.28 | 115.58 |
| Referent unit | 1 | useful/total | 43 | 0.35 | 0.06 |
| Referent unit | 1 | useful_token_count | 43 | 55.91 | 40.75 |
| Quantity | 0 | tokens_per_sentence | 28 | 12.98 | 7.08 |
| Quantity | 0 | total_token_count | 28 | 206.86 | 117.11 |
| Quantity | 0 | useful/total | 28 | 0.35 | 0.04 |
| Quantity | 0 | useful_token_count | 28 | 72.43 | 43.36 |
| Quantity | 1 | tokens_per_sentence | 32 | 12.55 | 7.24 |
| Quantity | 1 | total_token_count | 32 | 136.75 | 96.66 |
| Quantity | 1 | useful/total | 32 | 0.35 | 0.06 |
| Quantity | 1 | useful_token_count | 32 | 46.94 | 32.66 |
| Invariance/Constant | 0 | tokens_per_sentence | 18 | 12.83 | 8.32 |
| Invariance/Constant | 0 | total_token_count | 18 | 231.06 | 117.18 |
| Invariance/Constant | 0 | useful/total | 18 | 0.36 | 0.05 |
| Invariance/Constant | 0 | useful_token_count | 18 | 80.72 | 41.44 |
| Invariance/Constant | 1 | tokens_per_sentence | 42 | 12.72 | 6.63 |
| Invariance/Constant | 1 | total_token_count | 42 | 143.07 | 99.06 |
| Invariance/Constant | 1 | useful/total | 42 | 0.35 | 0.05 |
| Invariance/Constant | 1 | useful_token_count | 42 | 49.45 | 35.62 |
| Covariance | 0 | tokens_per_sentence | 19 | 14.46 | 9.97 |
| Covariance | 0 | total_token_count | 19 | 220.26 | 105.04 |
| Covariance | 0 | useful/total | 19 | 0.36 | 0.04 |
| Covariance | 0 | useful_token_count | 19 | 76.79 | 34.81 |
| Covariance | 1 | tokens_per_sentence | 41 | 11.96 | 5.27 |
| Covariance | 1 | total_token_count | 41 | 145.93 | 107.58 |
| Covariance | 1 | useful/total | 41 | 0.35 | 0.06 |
| Covariance | 1 | useful_token_count | 41 | 50.51 | 39.61 |

**Table S8 (continued).** Theme-Level Feature Characteristics

| Theme | Correct | Feature | N | Mean | SD |
|---|---|---|---|---|---|
| | | **Uninstructed Few-Shot-CoT** | | | |
| Referent unit | 0 | tokens_per_sentence | 17 | 9.61 | 2.30 |
| Referent unit | 0 | total_token_count | 17 | 180.24 | 99.09 |
| Referent unit | 0 | useful/total | 17 | 0.36 | 0.04 |
| Referent unit | 0 | useful_token_count | 17 | 63.47 | 35.67 |
| Referent unit | 1 | tokens_per_sentence | 43 | 13.99 | 7.96 |
| Referent unit | 1 | total_token_count | 43 | 165.21 | 116.85 |
| Referent unit | 1 | useful/total | 43 | 0.35 | 0.06 |
| Referent unit | 1 | useful_token_count | 43 | 57.00 | 41.59 |
| Quantity | 0 | tokens_per_sentence | 28 | 12.19 | 7.15 |
| Quantity | 0 | total_token_count | 28 | 189.82 | 116.19 |
| Quantity | 0 | useful/total | 28 | 0.36 | 0.04 |
| Quantity | 0 | useful_token_count | 28 | 66.93 | 41.43 |
| Quantity | 1 | tokens_per_sentence | 32 | 13.24 | 7.15 |
| Quantity | 1 | total_token_count | 32 | 151.66 | 105.80 |
| Quantity | 1 | useful/total | 32 | 0.35 | 0.06 |
| Quantity | 1 | useful_token_count | 32 | 51.75 | 37.56 |
| Invariance/Constant | 0 | tokens_per_sentence | 21 | 12.61 | 7.68 |
| Invariance/Constant | 0 | total_token_count | 21 | 222.76 | 116.31 |
| Invariance/Constant | 0 | useful/total | 21 | 0.37 | 0.05 |
| Invariance/Constant | 0 | useful_token_count | 21 | 79.43 | 40.18 |
| Invariance/Constant | 1 | tokens_per_sentence | 39 | 12.83 | 6.88 |
| Invariance/Constant | 1 | total_token_count | 39 | 140.77 | 98.80 |
| Invariance/Constant | 1 | useful/total | 39 | 0.34 | 0.05 |
| Invariance/Constant | 1 | useful_token_count | 39 | 47.74 | 35.37 |
| Covariance | 0 | tokens_per_sentence | 22 | 13.95 | 9.35 |
| Covariance | 0 | total_token_count | 22 | 220.55 | 106.00 |
| Covariance | 0 | useful/total | 22 | 0.35 | 0.05 |
| Covariance | 0 | useful_token_count | 22 | 76.68 | 36.04 |
| Covariance | 1 | tokens_per_sentence | 38 | 12.06 | 5.44 |
| Covariance | 1 | total_token_count | 38 | 139.89 | 104.84 |
| Covariance | 1 | useful/total | 38 | 0.35 | 0.06 |
| Covariance | 1 | useful_token_count | 38 | 48.50 | 38.62 |

**Table S8 (continued).** Theme-Level Feature Characteristics

| Theme | Correct | Feature | N | Mean | SD |
|---|---|---|---|---|---|
| | | **Uninstructed Few-shot-CoT-SC** | | | |
| Referent unit | 0 | tokens_per_sentence | 17 | 9.61 | 2.30 |
| Referent unit | 0 | total_token_count | 17 | 180.24 | 99.09 |
| Referent unit | 0 | useful/total | 17 | 0.36 | 0.04 |
| Referent unit | 0 | useful_token_count | 17 | 63.47 | 35.67 |
| Referent unit | 1 | tokens_per_sentence | 43 | 13.99 | 7.96 |
| Referent unit | 1 | total_token_count | 43 | 165.21 | 116.85 |
| Referent unit | 1 | useful/total | 43 | 0.35 | 0.06 |
| Referent unit | 1 | useful_token_count | 43 | 57.00 | 41.59 |
| Quantity | 0 | tokens_per_sentence | 26 | 12.60 | 7.27 |
| Quantity | 0 | total_token_count | 26 | 196.81 | 117.66 |
| Quantity | 0 | useful/total | 26 | 0.36 | 0.05 |
| Quantity | 0 | useful_token_count | 26 | 69.58 | 41.84 |
| Quantity | 1 | tokens_per_sentence | 34 | 12.87 | 7.09 |
| Quantity | 1 | total_token_count | 34 | 148.56 | 103.41 |
| Quantity | 1 | useful/total | 34 | 0.35 | 0.06 |
| Quantity | 1 | useful_token_count | 34 | 50.62 | 36.71 |
| Invariance/Constant | 0 | tokens_per_sentence | 20 | 12.75 | 7.86 |
| Invariance/Constant | 0 | total_token_count | 20 | 229.95 | 114.45 |
| Invariance/Constant | 0 | useful/total | 20 | 0.37 | 0.05 |
| Invariance/Constant | 0 | useful_token_count | 20 | 81.75 | 39.75 |
| Invariance/Constant | 1 | tokens_per_sentence | 40 | 12.75 | 6.81 |
| Invariance/Constant | 1 | total_token_count | 40 | 139.23 | 98.01 |
| Invariance/Constant | 1 | useful/total | 40 | 0.35 | 0.05 |
| Invariance/Constant | 1 | useful_token_count | 40 | 47.38 | 34.99 |
| Covariance | 0 | tokens_per_sentence | 22 | 13.95 | 9.35 |
| Covariance | 0 | total_token_count | 22 | 220.55 | 106.00 |
| Covariance | 0 | useful/total | 22 | 0.35 | 0.05 |
| Covariance | 0 | useful_token_count | 22 | 76.68 | 36.04 |
| Covariance | 1 | tokens_per_sentence | 38 | 12.06 | 5.44 |
| Covariance | 1 | total_token_count | 38 | 139.89 | 104.84 |
| Covariance | 1 | useful/total | 38 | 0.35 | 0.06 |
| Covariance | 1 | useful_token_count | 38 | 48.50 | 38.62 |

**Table S8 (continued).** Theme-Level Feature Characteristics

| Theme | Correct | Feature | N | Mean | SD |
|---|---|---|---|---|---|
| | | **Instructed Few-Shot** | | | |
| Referent unit | 0 | tokens_per_sentence | 15 | 10.23 | 2.69 |
| Referent unit | 0 | total_token_count | 15 | 180.93 | 109.36 |
| Referent unit | 0 | useful/total | 15 | 0.35 | 0.02 |
| Referent unit | 0 | useful_token_count | 15 | 62.00 | 37.43 |
| Referent unit | 1 | tokens_per_sentence | 45 | 13.59 | 7.91 |
| Referent unit | 1 | total_token_count | 45 | 165.64 | 113.14 |
| Referent unit | 1 | useful/total | 45 | 0.35 | 0.06 |
| Referent unit | 1 | useful_token_count | 45 | 57.78 | 40.93 |
| Quantity | 0 | tokens_per_sentence | 27 | 11.67 | 3.06 |
| Quantity | 0 | total_token_count | 27 | 220.78 | 119.42 |
| Quantity | 0 | useful/total | 27 | 0.35 | 0.04 |
| Quantity | 0 | useful_token_count | 27 | 76.70 | 43.67 |
| Quantity | 1 | tokens_per_sentence | 33 | 13.64 | 9.15 |
| Quantity | 1 | total_token_count | 33 | 127.48 | 85.37 |
| Quantity | 1 | useful/total | 33 | 0.35 | 0.06 |
| Quantity | 1 | useful_token_count | 33 | 44.21 | 29.73 |
| Invariance/Constant | 0 | tokens_per_sentence | 11 | 14.20 | 10.56 |
| Invariance/Constant | 0 | total_token_count | 11 | 209.82 | 122.87 |
| Invariance/Constant | 0 | useful/total | 11 | 0.35 | 0.06 |
| Invariance/Constant | 0 | useful_token_count | 11 | 72.27 | 45.18 |
| Invariance/Constant | 1 | tokens_per_sentence | 49 | 12.43 | 6.19 |
| Invariance/Constant | 1 | total_token_count | 49 | 160.41 | 108.05 |
| Invariance/Constant | 1 | useful/total | 49 | 0.35 | 0.05 |
| Invariance/Constant | 1 | useful_token_count | 49 | 55.82 | 38.37 |
| Covariance | 0 | tokens_per_sentence | 4 | 10.87 | 1.62 |
| Covariance | 0 | total_token_count | 4 | 273.00 | 116.52 |
| Covariance | 0 | useful/total | 4 | 0.36 | 0.04 |
| Covariance | 0 | useful_token_count | 4 | 94.50 | 33.51 |
| Covariance | 1 | tokens_per_sentence | 56 | 12.89 | 7.33 |
| Covariance | 1 | total_token_count | 56 | 162.07 | 108.43 |
| Covariance | 1 | useful/total | 56 | 0.35 | 0.05 |
| Covariance | 1 | useful_token_count | 56 | 56.29 | 39.25 |

**Table S8 (continued).** Theme-Level Feature Characteristics

| Theme | Correct | Feature | N | Mean | SD |
|---|---|---|---|---|---|
| | | **Instructed Few-Shot-SC** | | | |
| Referent unit | 0 | tokens_per_sentence | 16 | 10.31 | 2.62 |
| Referent unit | 0 | total_token_count | 16 | 183.31 | 106.08 |
| Referent unit | 0 | useful/total | 16 | 0.35 | 0.02 |
| Referent unit | 0 | useful_token_count | 16 | 63.00 | 36.38 |
| Referent unit | 1 | tokens_per_sentence | 44 | 13.64 | 8.00 |
| Referent unit | 1 | total_token_count | 44 | 164.43 | 114.15 |
| Referent unit | 1 | useful/total | 44 | 0.35 | 0.06 |
| Referent unit | 1 | useful_token_count | 44 | 57.32 | 41.29 |
| Quantity | 0 | tokens_per_sentence | 25 | 11.47 | 3.14 |
| Quantity | 0 | total_token_count | 25 | 207.44 | 117.38 |
| Quantity | 0 | useful/total | 25 | 0.35 | 0.04 |
| Quantity | 0 | useful_token_count | 25 | 72.84 | 43.82 |
| Quantity | 1 | tokens_per_sentence | 35 | 13.67 | 8.87 |
| Quantity | 1 | total_token_count | 35 | 142.34 | 100.12 |
| Quantity | 1 | useful/total | 35 | 0.35 | 0.06 |
| Quantity | 1 | useful_token_count | 35 | 48.83 | 33.88 |
| Invariance/Constant | 0 | tokens_per_sentence | 11 | 13.93 | 10.63 |
| Invariance/Constant | 0 | total_token_count | 11 | 237.45 | 136.77 |
| Invariance/Constant | 0 | useful/total | 11 | 0.35 | 0.06 |
| Invariance/Constant | 0 | useful_token_count | 11 | 81.27 | 48.71 |
| Invariance/Constant | 1 | tokens_per_sentence | 49 | 12.49 | 6.18 |
| Invariance/Constant | 1 | total_token_count | 49 | 154.20 | 100.41 |
| Invariance/Constant | 1 | useful/total | 49 | 0.35 | 0.05 |
| Invariance/Constant | 1 | useful_token_count | 49 | 53.80 | 36.24 |
| Covariance | 0 | tokens_per_sentence | 2 | 11.03 | 1.29 |
| Covariance | 0 | total_token_count | 2 | 325.00 | 155.56 |
| Covariance | 0 | useful/total | 2 | 0.39 | 0.08 |
| Covariance | 0 | useful_token_count | 2 | 119.00 | 33.94 |
| Covariance | 1 | tokens_per_sentence | 58 | 12.81 | 7.22 |
| Covariance | 1 | total_token_count | 58 | 164.10 | 107.53 |
| Covariance | 1 | useful/total | 58 | 0.35 | 0.05 |
| Covariance | 1 | useful_token_count | 58 | 56.76 | 38.59 |

**Table S8 (continued).** Theme-Level Feature Characteristics

| Theme | Correct | Feature | N | Mean | SD |
|---|---|---|---|---|---|
| | | **Instructed Few-Shot-CoT** | | | |
| Referent unit | 0 | tokens_per_sentence | 12 | 9.81 | 2.36 |
| Referent unit | 0 | total_token_count | 12 | 179.67 | 100.77 |
| Referent unit | 0 | useful/total | 12 | 0.35 | 0.02 |
| Referent unit | 0 | useful_token_count | 12 | 62.00 | 34.57 |
| Referent unit | 1 | tokens_per_sentence | 48 | 13.49 | 7.70 |
| Referent unit | 1 | total_token_count | 48 | 166.92 | 114.86 |
| Referent unit | 1 | useful/total | 48 | 0.35 | 0.06 |
| Referent unit | 1 | useful_token_count | 48 | 58.04 | 41.32 |
| Quantity | 0 | tokens_per_sentence | 20 | 13.72 | 8.16 |
| Quantity | 0 | total_token_count | 20 | 220.75 | 114.92 |
| Quantity | 0 | useful/total | 20 | 0.35 | 0.05 |
| Quantity | 0 | useful_token_count | 20 | 76.65 | 41.01 |
| Quantity | 1 | tokens_per_sentence | 40 | 12.27 | 6.58 |
| Quantity | 1 | total_token_count | 40 | 143.83 | 101.71 |
| Quantity | 1 | useful/total | 40 | 0.35 | 0.06 |
| Quantity | 1 | useful_token_count | 40 | 49.93 | 36.53 |
| Invariance/Constant | 0 | tokens_per_sentence | 9 | 14.43 | 11.79 |
| Invariance/Constant | 0 | total_token_count | 9 | 189.00 | 93.55 |
| Invariance/Constant | 0 | useful/total | 9 | 0.35 | 0.06 |
| Invariance/Constant | 0 | useful_token_count | 9 | 63.11 | 28.72 |
| Invariance/Constant | 1 | tokens_per_sentence | 51 | 12.45 | 6.06 |
| Invariance/Constant | 1 | total_token_count | 51 | 166.02 | 114.82 |
| Invariance/Constant | 1 | useful/total | 51 | 0.35 | 0.05 |
| Invariance/Constant | 1 | useful_token_count | 51 | 58.08 | 41.65 |
| Covariance | 0 | tokens_per_sentence | 7 | 17.18 | 12.96 |
| Covariance | 0 | total_token_count | 7 | 146.14 | 85.14 |
| Covariance | 0 | useful/total | 7 | 0.33 | 0.05 |
| Covariance | 0 | useful_token_count | 7 | 46.57 | 24.34 |
| Covariance | 1 | tokens_per_sentence | 53 | 12.17 | 5.91 |
| Covariance | 1 | total_token_count | 53 | 172.55 | 114.81 |
| Covariance | 1 | useful/total | 53 | 0.36 | 0.05 |
| Covariance | 1 | useful_token_count | 53 | 60.45 | 41.32 |

**Table S8 (continued).** Theme-Level Feature Characteristics

| Theme | Correct | Feature | N | Mean | SD |
|---|---|---|---|---|---|
| | | **Instructed Few-shot-CoT-SC** | | | |
| Referent unit | 0 | tokens_per_sentence | 13 | 9.81 | 2.34 |
| Referent unit | 0 | total_token_count | 13 | 196.38 | 107.80 |
| Referent unit | 0 | useful/total | 13 | 0.35 | 0.02 |
| Referent unit | 0 | useful_token_count | 13 | 67.46 | 36.84 |
| Referent unit | 1 | tokens_per_sentence | 47 | 13.56 | 7.76 |
| Referent unit | 1 | total_token_count | 47 | 162.02 | 112.46 |
| Referent unit | 1 | useful/total | 47 | 0.35 | 0.06 |
| Referent unit | 1 | useful_token_count | 47 | 56.45 | 40.65 |
| Quantity | 0 | tokens_per_sentence | 19 | 13.85 | 8.36 |
| Quantity | 0 | total_token_count | 19 | 225.32 | 116.19 |
| Quantity | 0 | useful/total | 19 | 0.35 | 0.05 |
| Quantity | 0 | useful_token_count | 19 | 78.00 | 41.67 |
| Quantity | 1 | tokens_per_sentence | 41 | 12.24 | 6.50 |
| Quantity | 1 | total_token_count | 41 | 143.59 | 100.44 |
| Quantity | 1 | useful/total | 41 | 0.35 | 0.06 |
| Quantity | 1 | useful_token_count | 41 | 49.95 | 36.08 |
| Invariance/Constant | 0 | tokens_per_sentence | 10 | 14.74 | 11.16 |
| Invariance/Constant | 0 | total_token_count | 10 | 177.10 | 95.89 |
| Invariance/Constant | 0 | useful/total | 10 | 0.35 | 0.06 |
| Invariance/Constant | 0 | useful_token_count | 10 | 59.30 | 29.64 |
| Invariance/Constant | 1 | tokens_per_sentence | 50 | 12.35 | 6.08 |
| Invariance/Constant | 1 | total_token_count | 50 | 167.94 | 115.16 |
| Invariance/Constant | 1 | useful/total | 50 | 0.35 | 0.05 |
| Invariance/Constant | 1 | useful_token_count | 50 | 58.74 | 41.80 |
| Covariance | 0 | tokens_per_sentence | 4 | 21.16 | 16.50 |
| Covariance | 0 | total_token_count | 4 | 161.00 | 100.76 |
| Covariance | 0 | useful/total | 4 | 0.31 | 0.03 |
| Covariance | 0 | useful_token_count | 4 | 51.25 | 33.88 |
| Covariance | 1 | tokens_per_sentence | 56 | 12.15 | 5.82 |
| Covariance | 1 | total_token_count | 56 | 170.07 | 113.02 |
| Covariance | 1 | useful/total | 56 | 0.35 | 0.05 |
| Covariance | 1 | useful_token_count | 56 | 59.38 | 40.42 |

## Supplemental Figures

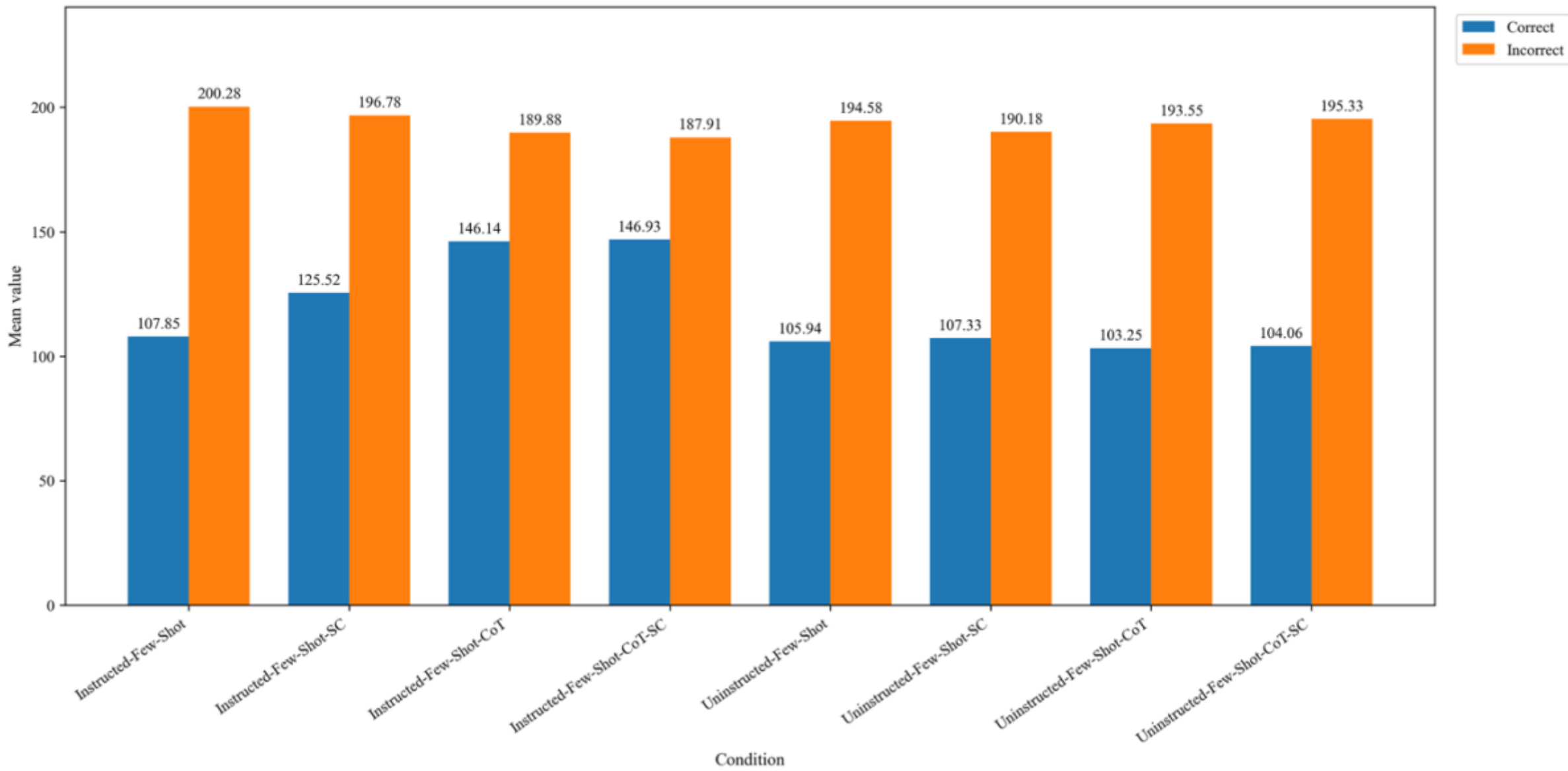

**Figure S1.** Total Token Count Comparison in Correct and Incorrect Cases

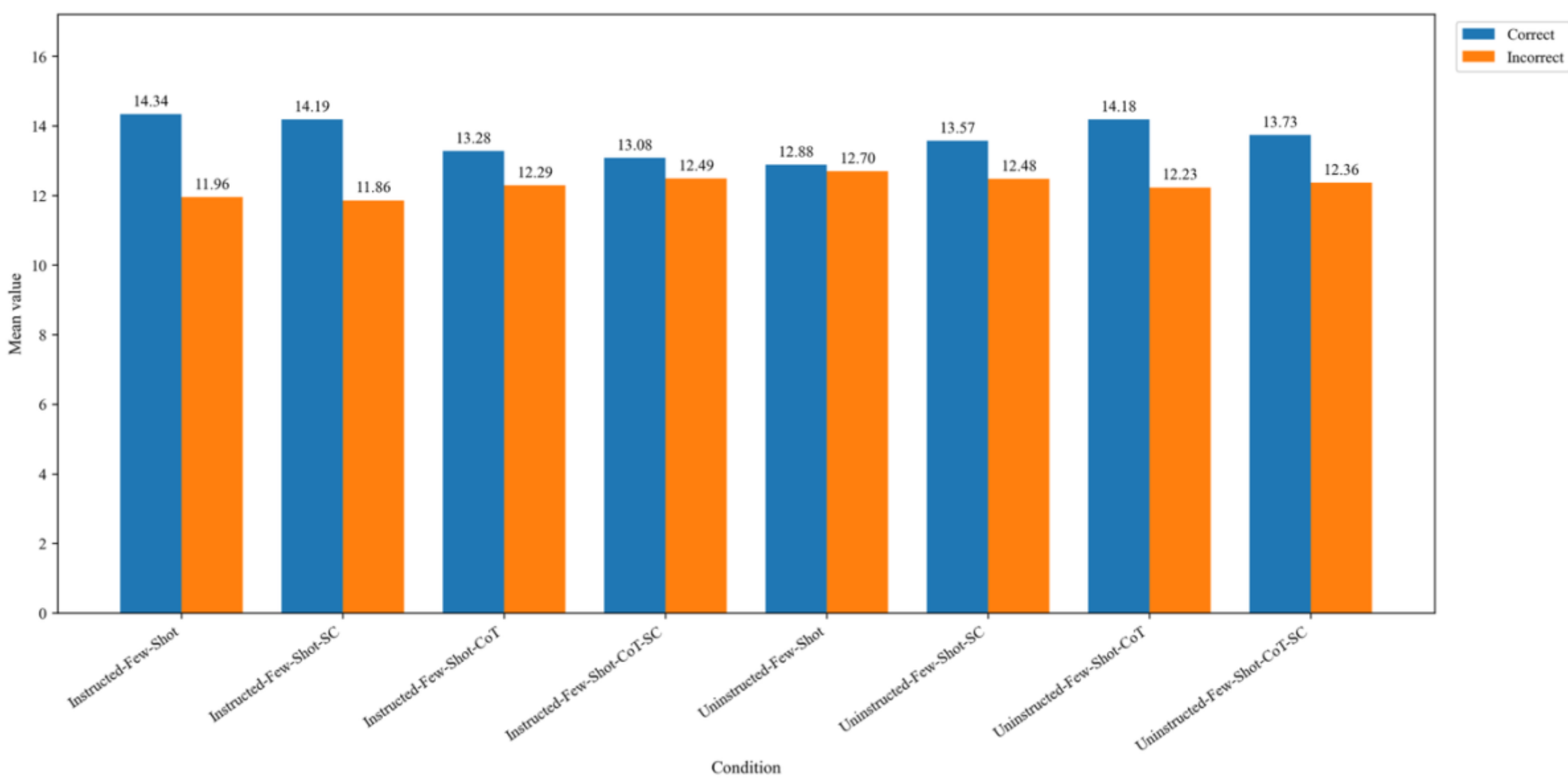

**Figure S2.** Tokens per Sentence Comparison in Correct and Incorrect Cases

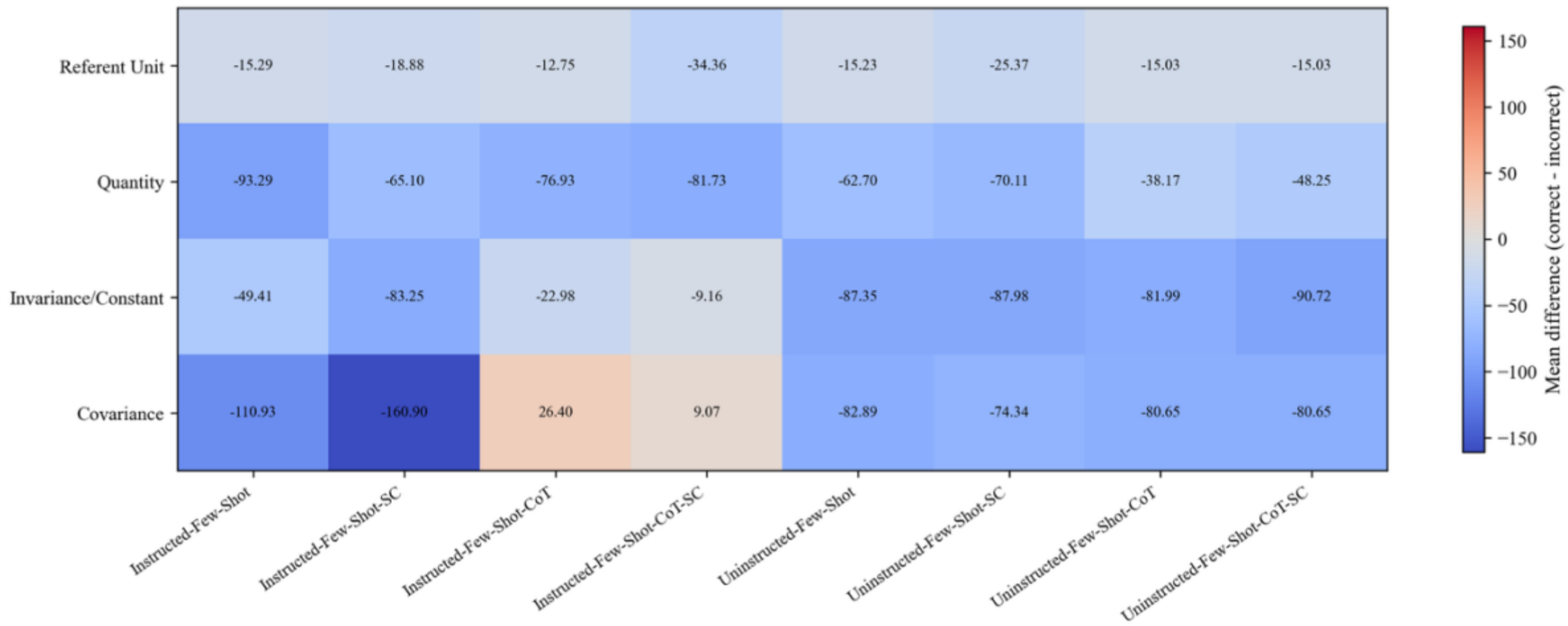

**Figure S3.** Mean Difference in Total Token Count Between Correct and Incorrect Cases at the Theme Level